\documentclass[conference]{ieeetran}  % Comment this line out if you need a4paper

\IEEEoverridecommandlockouts                              % This command is only needed if 
\usepackage{graphicx}   % For including images (\includegraphics)
\usepackage{cite}       % For formatting citations (e.g., [1]-[3])
\usepackage{amsmath}    % For advanced math environments (split, multlined, etc.)
\usepackage{amssymb}    % For additional math symbols (e.g., \mathbb{E})
\usepackage{amsfonts}   % For math fonts
\usepackage{subcaption}
\usepackage{algorithm}      % For floating algorithm environments
\usepackage{algpseudocode}  % For pseudocode within algorithm
\usepackage{listings}       % For code listings with syntax highlighting and line breaking
\usepackage{multirow}
\usepackage{booktabs}   % For professional-quality tables (\toprule, \midrule)
\usepackage[most]{tcolorbox}  % For creating styled text boxes (like our promptbox)
\usepackage{xcolor}     % For using colors (loaded by tcolorbox, but good to be explicit)

\usepackage{hyperref}   % For clickable links (citations, URLs, references)

\newtcolorbox{promptbox}[2][]{ % The '2' means this environment takes 2 arguments
    enhanced, % Allows for more advanced features
    attach boxed title to top left={yshift=-2mm, xshift=4mm},
    colback=blue!5!white,
    colframe=blue!75!black,
    colbacktitle=blue!75!black,
    coltitle=white,
    fonttitle=\bfseries,
    title=#2, % The second argument is used as the title
    arc=4mm,
    boxrule=1pt,
    #1 % The first (optional) argument is for any other options
}
\def\BibTeX{{\rm B\kern-.05em{\sc i\kern-.025em b}\kern-.08em
    T\kern-.1667em\lower.7ex\hbox{E}\kern-.125emX}}
\begin{document}

\title{Reinforced Embodied Planning with Verifiable Reward for Real-World Robotic Manipulation\\
% {\footnotesize \textsuperscript{*}Note: Sub-titles are not captured in Xplore and
% should not be used}
% \thanks{Identify applicable funding agency here. If none, delete this.}
}
\author{
Zitong Bo\thanks{*Equal contribution.}\textsuperscript{\dag},\
Yue Hu*\textsuperscript{\dag\ddag},\
Jinming Ma\textsuperscript{\dag},\
Mingliang Zhou\textsuperscript{\dag},\
Junhui Yin\textsuperscript{\dag},\
Yachen Kang\textsuperscript{\dag},\
Yuqi Liu\textsuperscript{\dag},\
Tong Wu\textsuperscript{\dag},\
Diyun Xiang\textsuperscript{\dag},\\  % <— 换行
and\ Hao Zhou\textsuperscript{\ddag}
\\[2mm]
\textsuperscript{\dag}\textit{Xiaomi Robotics Lab}\\
\textsuperscript{\ddag}\textit{College of Computer Science and Technology, Zhejiang University}\\
\textit{Corresponding author: bozitong@xiaomi.com}
}

\maketitle

\begin{abstract}
Enabling robots to execute long-horizon manipulation tasks from free-form language instructions remains a fundamental challenge in embodied AI. While vision–language models (VLMs) have shown promise as high-level planners, their deployment in the real world is hindered by two gaps: (i) the scarcity of large-scale, sequential manipulation data that couples natural language with multi-step action plans, and (ii) the absence of dense, interpretable rewards for fine-tuning VLMs on planning objectives. 
To address these issues, we propose \textbf{REVER}, a framework that empowers VLMs to generate and validate long-horizon manipulation plans from natural language instructions in real-world scenarios. Under REVER we train and release \textbf{RoboFarseer}, a VLM incentivized to emit chain-of-thought that perform temporal and spatial reasoning, ensuring physically plausible and logically coherent plans. To obtain training data, we leverage the Universal Manipulation Interface framework to capture hardware-agnostic demonstrations of atomic skills. An automated annotation engine converts each demonstration into vision–instruction–plan triplet. We introduce a verifiable reward that scores the generated plan by its ordered bipartite matching overlap with the ground-truth skill sequence. At run time, the fine-tuned VLM functions both as a planner and as a monitor, verifying step-wise completion. 
RoboFarseer matches or exceeds the performance of proprietary models that are orders of magnitude larger, while on open-ended planning it surpasses the best baseline by more than 40\%. In real-world, long-horizon tasks, the complete system boosts overall success by roughly 60\% compared with the same low-level controller without the planner. We will open-source both the dataset and the trained model upon publication.
\end{abstract}


\section{Introduction}
Embodied planning is a cornerstone of intelligent robotics: an agent must perceive a dynamic environment, reason about latent goals, and produce coherent, long-horizon action sequences that satisfy free-form natural-language instructions~\cite{wu2023embodied,shi2025hirobot}. Whereas low-level controllers concentrate on accurate trajectory tracking~\cite{chi2023dp,zitkovich2023rt}, high-level planning operates at the intersection of vision, language, and action, demanding temporally extended, physically grounded decision-making. Recent large vision–language models (VLMs) unify visual perception and linguistic reasoning in a single backbone~\cite{bordes2024introduction,huang2025vlm}, yet turning them into active, reliable planners remains elusive. In open-world manipulation, even the strongest VLMs hallucinate unreachable states, violate physical constraints, or derail after the first unexpected observation~\cite{wu2025reinforced}, underscoring a gap between static reasoning and the incremental, feedback-rich nature of real-world task execution.

A fundamental impediment is the learning paradigm itself. Supervised fine-tuning (SFT) on human-collected plan traces teaches VLMs to imitate expert sequences, but provides no mechanism for recovery or creative recombination once the context drifts. Reinforcement learning (RL) promises adaptive, self-improving behaviour, yet embodied RL for VLMs is still nascent: prior work largely targets spatial-reasoning\cite{yuan2025embodiedr1,azzolini2025cosmos,ji2025robobrain} or robotic simulator where reward engineering is trivial. When the objective is open-ended planning, dense, interpretable rewards are notoriously difficult to craft. LLM-as-a-Judge methods~\cite{whitehouse2025j1} supply noisy, unverifiable scalar signals that vary unpredictably across prompts, whereas hand-crafted heuristics break down as scene complexity grows. Compounding the problem is data scarcity: existing real-world collections such as Universal Manipulation Interface (UMI)~\cite{chi2024umi} excel at short, atomic skills but lack the compositional breadth and sequential annotation needed to train and evaluate long-horizon planners. Consequently, existing VLM planners either remain confined to synthetic benchmarks with canned instructions or degrade rapidly when confronted with the visual diversity, action variability, and partial observability of real households.

\begin{figure}
    \centering
    \includegraphics[width=0.9\linewidth,trim=2 2 2 2,clip]{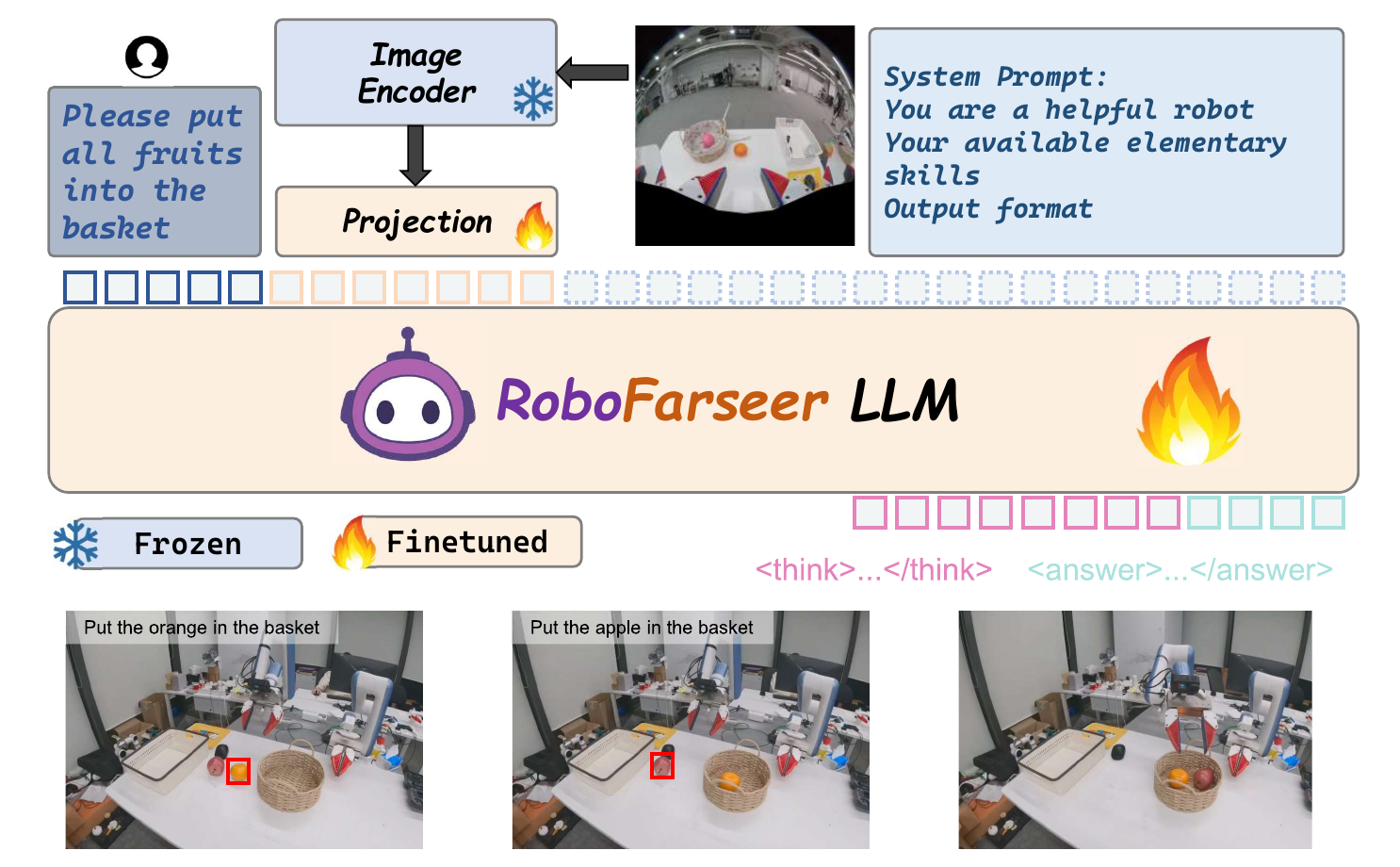}
    \caption{Overview of \textbf{RoboFarseer}.}
    \label{fig:placeholder}
    \vspace{-7mm}
\end{figure}
We bridge these gaps with \textbf{REVER} (\textbf{Re}inforced \textbf{E}mbodied planning with \textbf{V}erifiable \textbf{R}eward)—a framework that turns a VLM into a reliable long-horizon planner and verifier for real-world manipulation.
Inspired by the DeepSeek-R1 paradigm~\cite{guo2025deepseek}, REVER incentivizes the VLM to produce an explicit chain-of-thought that performs temporal and spatial reasoning before committing to any skill, ensuring the ensuing plan is both physically plausible and logically coherent.
Within this pipeline we train and release \textbf{RoboFarseer},  fine-tuned by Qwen2.5-VL-7B\cite{bai2025qwen2}, that learns to generate and self-evaluate such thought-augmented plans.
Central to REVER is a deterministic reward that inspects a proposed plan along two axes: (1) syntactic validity—every step must instantiate a predefined, executable skill grammar—and (2) semantic coverage—an ordered bipartite match against the ground-truth skill sequence.
This reward is interpretable, and cheap to compute, allowing us to formulate fine-tuning as Reinforcement Learning with Verifiable Reward (RLVR) without human labels or simulator queries.
During RLVR, RoboFarseer learns two complementary capabilities in a single forward pass: emitting an instruction-conditioned multi-step plan and estimating whether the current skill has been completed, yielding a closed-loop system that advances autonomously through extended tasks.
To train and evaluate RoboFarseer, we curate a large-scale, sequential dataset.
We extend UMI framework with an automated pipeline that converts in-the-wild, kinesthetic video demonstrations into Vision-Instruction-Plan triplets.
The pipeline segments videos into atomic skills, labels them with language descriptions, and assembles composite plans.
Our contributions are summarized as follow:
\begin{itemize}
\item We propose \textbf{REVER}, a reinforcement-learning framework that leverages a \emph{verifiable reward} derived from skill-syntax checks and bipartite matching, and performs reinforcement fine-tuning to yield \textbf{RoboFarseer}.
\item We introduce an automated pipeline that converts raw UMI demonstrations into Vision–Instruction–Plan triplets. Building on this pipeline, we present the \textbf{LEAP} (\textbf{L}ong-horizon \textbf{E}mbodied \textbf{A}ction \textbf{P}lanning) dataset, which contains two subsets—\textbf{LEAP-L} for long-horizon sequential planning and \textbf{LEAP-U} for instruction-aligned planning.
\item We demonstrate a closed-loop, hierarchical system in which \textbf{RoboFarseer} simultaneously \emph{generates} high-level skill plans and \emph{monitors} their execution, validated on public benchmarks and through real-robot deployments on complex, long-horizon household tasks.
\end{itemize}

\S\ref{sec:related_work} reviews related work on embodied planning, visual reasoning, and real-world robotic datasets. \S\ref{sec:method} details the REVER framework, including data synthesis, verifiable-reward design, and hierarchical execution. \S\ref{sec:experiment} presents benchmarks and real-robot evaluations, and \S\ref{sec:conclusion} concludes with future directions.

All model weights, datasets, and source code will be made publicly available upon acceptance to facilitate reproducibility.
\section{Related Work}
\label{sec:related_work}

\subsection{Embodied Planning}
Embodied planning aims to decompose high-level natural language instructions into executable sub-task sequences, enabling embodied agents to perform complex behaviors
in interactive environments~\cite{shin2024socratic,fei2025unleashing}. The advent of VLMs has spurred their use as high-level planners~\cite{wu2023embodied,shi2025hirobot}. Initial efforts utilized few-shot prompting with frozen VLMs to generate plans~\cite{wu2023embodied}, but they often faced challenges with spatial grounding and temporal coherence. To improve performance, a dominant paradigm is to fine-tune VLMs on demonstration data, often in a hierarchical structure~\cite{driess2023palme,zhang2025embodiedreasoner}. In this setup, a VLM planner generates interpretable sub-tasks that are then executed by a low-level policy~\cite{shi2025hirobot,wen2025dexvla}. A significant portion of this research, however, relies heavily on synthetic environments\cite{cai2025cookbench,yang2025embodiedbench} or operates with fixed, predefined action and instruction sets~\cite{shao2025large}. Our work focuses on open-ended planning and execution directly in the physical world.

\subsection{Visual Reasoning}
To elicit structured reasoning, prior works teach VLMs to generate explicit Chain-of-Thought (CoT) via supervised fine-tuning (SFT) on step-by-step demonstrations~\cite{chen2023robogpt,zhang2025embodiedreasoner}.
While effective, SFT is bounded by demonstration diversity.
Reinforcement Fine-Tuning (RFT) further improves reasoning by optimizing human preferences~\cite{rafailov2023dpo} or rule-based rewards~\cite{shao2024deepseekmath}.
Embodied foundation models increasingly adopt RFT to improve temporal reasoning~\cite{yuan2025fsd,ji2025robobrain,team2025robobrain2,azzolini2025cosmos}; however, they still require concrete instructions (e.g., "put the pot on the blue cloth") that produce visual guidance, rather than interpreting implicit human requests for long-horizon planning.
In contrast, our approach, we target long-horizon planning reasoning and devise a \emph{verifiable} reward grounded in skill grammars and bipartite matching, enabling RL fine-tuning without human preference labels\cite{rafailov2023dpo} or LLM-as-a-Judge\cite{whitehouse2025j1}.

\subsection{Real-World Data for Robotic Learning}
Bridging the sim-to-real gap demands large-scale, diverse, real-world data.
RT-1 and RT-2~\cite{brohan2022rt1,zitkovich2023rt} showed that combining internet-scale pre-training with real robot trajectories improves generalization.
The UMI framework~\cite{chi2024umi} enables hardware-agnostic, kinesthetic demonstrations in the wild.
Recent efforts such as Open-X-Embodiment~\cite{o2024oxe} aggregate multi-robot trajectories to scale up real-world data diversity, yet they mainly focus on short-horizon action policy learning. Meanwhile, datasets like ShareRobot\cite{ji2025robobrain} require labor-intensive manual annotation to obtain language labels.
We build upon UMI by automatically structuring its demonstrations into Vision–Instruction–Plan triplets and releasing the LEAP dataset for long-horizon planning research.

\section{Methodology}
\label{sec:method}
\begin{figure*}[ht!]
    \centering
    \includegraphics[width=\linewidth]{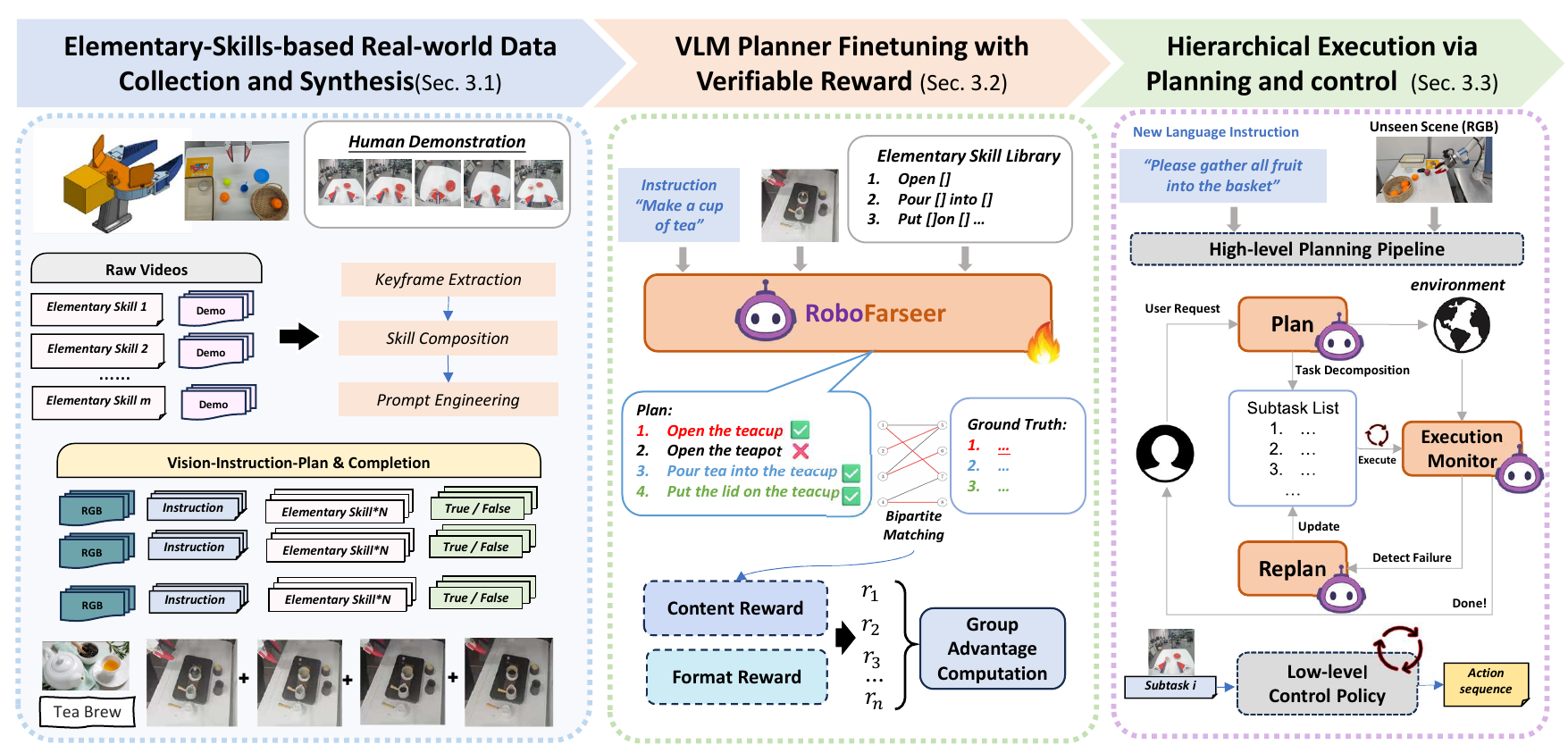}
    \caption{An overview of REVER framework.}
    \label{fig:framework}
    \vspace{-4mm}
\end{figure*}
As shown in Fig.~\ref{fig:framework}, REVER comprises three core components: (1) a systematic data collection and synthesis pipeline that converts real-world demonstrations into a structured format for VLM training; (2) a reinforcement fine-tuning method with verifiable reward for incentivizing planning ability of VLM; and (3) a hierarchical execution framework combining high-level planning and low-level control for real-world deployment. 

\subsection{Elementary-Skills-based Real-world Data Collection and Synthesis}
\label{ssec:data_pipeline}

A foundational requirement for our work is a dataset that captures long-horizon, language-conditioned manipulation in the real world. We develop a systematic data generation pipeline to process raw, in-the-wild video demonstrations into a structured, multi-purpose training corpus. The pipeline consists of three main stages: skill acquisition, composition, and automated annotation generation.

\subsubsection{Skill Acquisition and Compositional Task Synthesis}
\begin{figure}[htbp!]
    \centering
    \includegraphics[width=\linewidth,trim=2 2 2 2,clip]{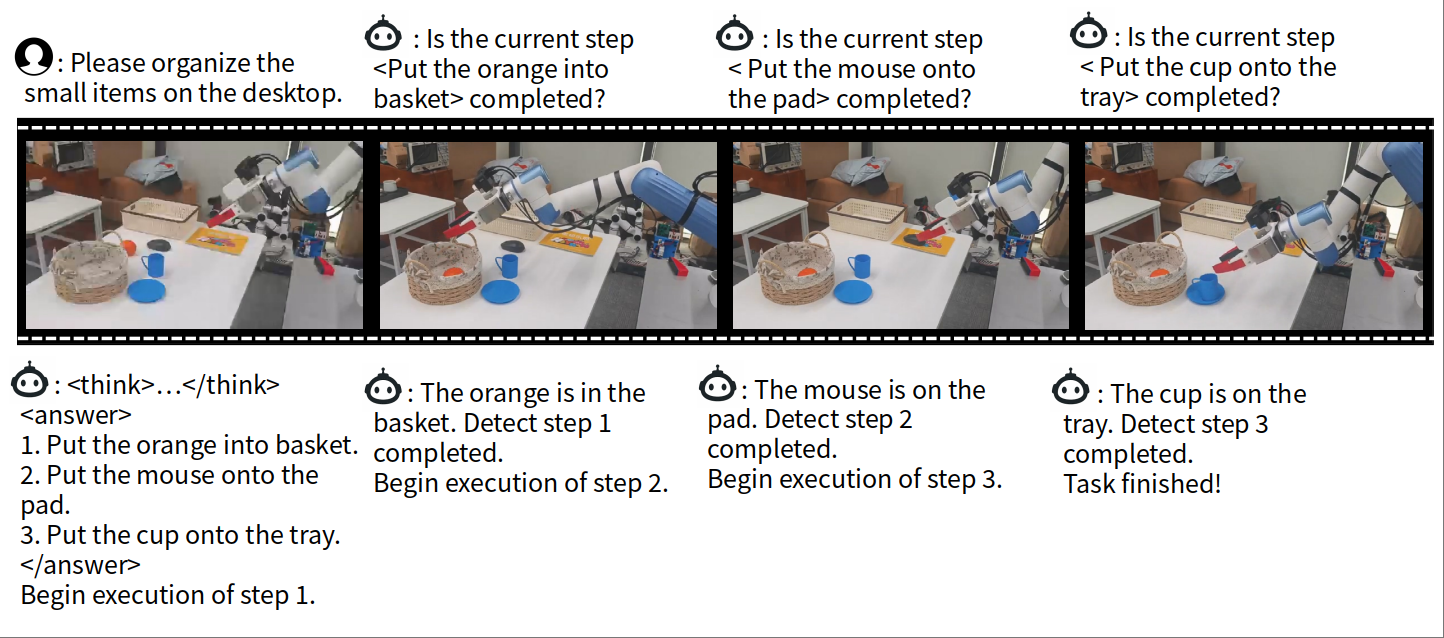}
    \caption{An example execution trajectory of RoboFarseer.}
    \label{fig:execution}
    \vspace{-4mm}
\end{figure}

We first collect a set of $N$ atomic, kinesthetically-rich manipulation skills, $\mathcal{S} = \{s_1, s_2, \ldots, s_N\}$, using the UMI gripper~\cite{chi2024umi}. Each skill $s_i$ is demonstrated multiple times, yielding a collection of raw video trajectories. A long-horizon task $\mathcal{T}$ is then programmatically synthesized by composing an ordered sequence of $K$ skills: $\mathcal{T} = (s^{(1)}, s^{(2)}, \ldots, s^{(K)})$, where each $s^{(k)} \in \mathcal{S}$. This compositional approach enables the creation of a vast and diverse set of complex tasks from a finite library of reusable skills. Fig.~\ref{fig:execution} shows an example of the long-horizon task "Organize the small items on the desktop", and more details about long-horizon tasks can be found in \S\ref{real-world}.

\subsubsection{Automated Annotation Generation}
 For each synthesized task $\mathcal{T}$, the pipeline  processes the corresponding video segments to generate annotations for two distinct learning objectives: planning and completion verification. For each sub-task $s^{(k)}$, the pipeline extracts keyframes. This includes an initial observation $o^{(k)}_{\text{init}}$ (taken before the sub-task begins) and a final observation $o^{(k)}_{\text{final}}$ (taken after). The pipeline then generates the following annotation types:
\begin{itemize}
    \item \textbf{Plan Annotation ($y_{\text{plan}}$)}: Given the initial observation $o^{(1)}_{\text{init}}$ and a high-level user instruction $I_{\text{user}}$, the ground-truth label is the full sequence of sub-tasks, i.e., $y_{\text{plan}} = (s^{(1)}, \ldots, s^{(K)})$.
    \item \textbf{Completion Annotation ($y_{\text{comp}}$)}: This is a binary classification task. Given the pair of observations $(o^{(k)}_{\text{init}}, o^{(k)}_{\text{final}})$ and the instruction for sub-task $s^{(k)}$, the label $y_{\text{comp}} \in \{\texttt{True}, \texttt{False}\}$ indicates whether the sub-task was successfully completed. We generate both positive samples from successful executions and negative samples by pairing $o^{(k)}_{\text{init}}$ with a frame $o'_{\text{uncomp}}$ sampled from the middle of the execution.
\end{itemize}

Finally, the pipeline assembles these annotations into a structured format. Each data point is a triplet $d = (q, \mathcal{O}, y)$, where $\mathcal{O}$ is a set of one or more observations, $q$ is the question prompt, and $y$ is the corresponding ground-truth answer from $\{y_{\text{plan}}, y_{\text{comp}}\}$. The prompts are dynamically generated from templates which include role-playing instructions, the available skill set $\mathcal{S}$ and output format. The resulting \textbf{LEAP} dataset contains 12,000 tasks and is split 9:1 for training and testing.

\subsection{VLM Planner fine-tuning with Verifiable Reward}
\label{ssec:reward_and_rl}

The cornerstone of REVER is our novel, verifiable reward function. This function provides a deterministic and interpretable signal for optimizing the planner's output using GRPO~\cite{shao2024deepseekmath}.

\subsubsection{The REVER Verifiable Reward Function}
Given a generated plan $P_g = (p_1, \ldots, p_M)$ and a ground-truth plan $P_{gt} = (p'_1, \ldots, p'_N)$, our reward function $\mathcal{R}(P_g, P_{gt})$ computes a score based on syntactic correctness, semantic alignment, and a penalty for length mismatch.
\begin{align}
\label{eq:total_reward}
\mathcal{R}(P_g, P_{gt}) = w_f \cdot \mathcal{R}_{\text{format}}(P_g) + w_c \cdot \mathcal{R}_{\text{content}}(P_g, P_{gt}) \notag
\end{align}
\paragraph{Format Score ($\mathcal{R}_{\text{format}}$)}
To ensure that the fine-tuned model consistently generates structured reasoning traces, we augment the verifiable reward with a \textbf{format reward} that penalizes any deviation from the required tag template. Formally, for a response $y$ we define
\[
R_{\text{format}}(P_g)=
\begin{cases}
1 & \parbox[t]{0.85\linewidth}{
      if $P_g$ matches
      $\mathtt{<think>\cdots</think>}$\\[2pt]
      $\mathtt{<answer>\cdots</answer>}$,
    }\\[10pt]
0 & \text{otherwise}.
\end{cases}
\]

\paragraph{Content Score ($\mathcal{R}_{\text{content}}$)}
This component measures the semantic similarity between the generated plan $P_g$ and the ground-truth plan $P_{gt}$. We model this as a maximum weight bipartite matching problem. We construct a bipartite graph where the two sets of vertices represent the steps of $P_g$ and $P_{gt}$. The weight of the edge between step $p_i \in P_g$ and $p'_j \in P_{gt}$ is given by a similarity function $\text{Sim}(p_i, p'_j)$. This function is a weighted sum of action similarity and object similarity:
\begin{equation}
\text{Sim}(p_i, p'_j) = w_a \cdot \text{Sim}_{\text{act}}(p_i, p'_j) + w_o \cdot \text{Sim}_{\text{obj}}(p_i, p'_j) \notag
\end{equation}
where $w_a=0.3$ and $w_o=0.7$. $\text{Sim}_{\text{act}}$ is 1 if the action verbs match (e.g., 'Put', 'Pick up') and 0 otherwise. $\text{Sim}_{\text{obj}}$ is 1 if the object/location arguments are deemed similar (i.e., they are an exact match, one is a substring of the other, or both belong to the same predefined semantic set from our ontology $\mathcal{C}$), and 0 otherwise. The total weight of the maximum matching, normalized by the length of the longer plan, is:
\begin{equation}
 \mathcal{R}_{\text{bm}}(P_g, P_{gt})=\frac{1}{\max(M, N)} \max_{\pi \in \Pi} \sum_{i=1}^{M} \text{Sim}(p_i, p'_{\pi(i)}) \notag
\end{equation}
where $\Pi$ is the set of all possible matchings. This score robustly handles reordered but semantically correct steps. Besides, we employ a length penalty, the content reward is:
\begin{equation}
    \mathcal{R}_{\text{content}}(P_g, P_{gt})=  \mathcal{R}_{\text{bm}}(P_g, P_{gt}) - w_l  \cdot | M - N | \notag
\end{equation}
where $w_l=0.1$. For completion data, the content reward $R_{content}$ is 1 when the output exactly matches the label and 0 otherwise.
\subsubsection{Planner Fine-tuning with GRPO}
Leveraging the proposed verifiable reward, we refine the VLM planner with GRPO~\cite{shao2024deepseekmath}. The algorithm then computes the relative advantage $\hat{A}_i$ for each response $P_i$ by normalizing its reward against the group's statistics. This is calculated as the standardized score of the reward:
\begin{equation}
\label{eq:advantage}
\hat{A}_i = \frac{r_i - \operatorname{mean}(\{r_1,\dots,r_B\})}{\operatorname{std}(\{r_1,\dots,r_B\})} \notag
\end{equation}

GRPO then optimizes a clipped surrogate objective function, a technique inspired by proximal policy optimization (PPO)~\cite{schulman2017ppo} to ensure stable policy updates. This objective aims to maximize the advantage of sampled actions, but penalizes large deviations from the reference policy $\pi_{\text{ref}}$. The objective is:
\begin{align}
\mathcal{J}_{\text{GRPO}}(\theta) &= \mathbb{E}_{\{o_i\}_{i=1}^N \sim \pi_{\theta_{\text{old}}}(q)} \notag \\
&[\frac{1}{N} \sum_{i=1}^{N} \left\{ \min[s_1 \cdot A_i,\; s_2 \cdot A_i] - \beta \mathbb{D}_{KL}[\pi_\theta || \pi_{ref}] \right\}] \notag \\
s_1 &= \frac{\pi_\theta(o_i|q)}{\pi_{\theta_{\text{old}}}(o_i|q)}, s_2 = \text{clip}\left(\frac{\pi_\theta(o_i|q)}{\pi_{\theta_{\text{old}}}(o_i|q)}, 1-\epsilon, 1+\epsilon\right) \notag
\end{align}

The $\min$ operator takes the pessimistic bound between the normal policy objective and a clipped version, where the probability ratio is restricted to the range $[1-\varepsilon, 1+\varepsilon]$. This clipping mechanism discourages excessively large updates that could destabilize the learning process. The KL-divergence term serves as an additional soft penalty to keep the learned policy from deviating too far from the trusted reference policy. By minimizing this loss, we encourage the policy to produce plans with higher advantages in a controlled and stable manner.

\subsection{Hierarchical Execution Framework via Planning and Control}
\label{ssec:hierarchical_framework}

To deploy our system onto a robot, we design a hierarchical execution framework that integrates our high-level VLM planner with a low-level control policy. This framework, detailed in Algorithm~\ref{alg:execution_framework}, enables the robot to autonomously handle long-horizon tasks through a dynamic loop of planning, execution, and monitoring.

\subsubsection{System Components}
\begin{itemize}
    \item \textbf{Planner}: RoboFarseer receives a high-level, semantically abstract user request and an initial visual observation of the environment to produce a structured, multi-step sub-task list.
    \item \textbf{Control Policy}: A low-level control policy that acts as the robot's action generator. It takes a single, concrete sub-task instruction (e.g., "Put the cup on the table"), along with current visual observations and robot proprioception data, to generate and execute low-level action.
    \item \textbf{Execution Monitor}: RoboFarseer operates in a verification mode. It continuously assesses whether the current sub-task has been completed, providing the critical feedback signal that drives the entire closed-loop system.
\end{itemize}

\subsubsection{Execution Flow}
The process begins when a user provides a high-level request. As detailed in Algorithm~\ref{alg:execution_framework}, the VLM Planner first performs planning, generating an initial sub-task list $P$. The system then enters the main execution loop, managed by the Execution Monitor. The monitor passes the current sub-task $s_{\text{current}}$ to the control policy. The control policy then begins its execution cycle. Crucially, while the control  policy is active, the VLM monitor continuously checks for task completion at a fixed frequency of 5Hz. It repeatedly takes the current visual observation $o_{\text{current}}$ and compares it against the initial state $o^{(k)}_{\text{prev}}$ to determine if the goal has been reached. Once the monitor outputs \texttt{True}, it signals the control policy to stop, and the system advances to the next sub-task in the list. If the system fails to achieve the goal within a certain time or number of attempts, the monitor feeds the current world state back to the VLM Planner, which generates an updated sub-task list to adapt to the new situation. This cycle continues until all sub-tasks are successfully completed. This architecture allows the robot to robustly execute complex plans while dynamically adapting to execution outcomes.

% CORRECTED Algorithm
\begin{algorithm}[htbp]
\caption{Hierarchical Planning and Execution Framework}
\label{alg:execution_framework}
\begin{algorithmic}[1]
\State \textbf{Input:} User request $I_{\text{user}}$, VLM planner $\pi_{\text{planner}}$, low-level control policy $\pi_{\text{control}}$
\State \textbf{Initialize:} Sub-task list $P \leftarrow \emptyset$, Task pointer $k \leftarrow 1$

\vspace{0.5em}
\Statex \textit{// -- Stage 1: Initial Planning --}
\State $o_{\text{init}} \leftarrow \text{CaptureInitialObservation}()$
\State $P = (s^{(1)}, \ldots, s^{(K)}) \leftarrow \pi_{\text{planner}}.\text{Plan}(I_{\text{user}}, o_{\text{init}})$
\If{$P$ is empty} \Return \texttt{Failure} \EndIf
\State $o^{(1)}_{\text{prev}} \leftarrow o_{\text{init}}$

\vspace{0.5em}
\Statex \textit{// -- Stage 2: Monitored Execution Loop --}
\While{$k \le K$}
    \State $s_{\text{current}} \leftarrow P[k]$
    \State \texttt{execution\_successful} $\leftarrow$ \Call{MonitoredExecute}{$s_{\text{current}}, o^{(k)}_{\text{prev}}$}
    \If{\texttt{execution\_successful}}
        \State $k \leftarrow k + 1$
        \State $o^{(k)}_{\text{prev}} \leftarrow \text{CaptureCurrentObservation}()$ 
    \Else
        \State \textit{// Re-plan due to execution failure}
        \State $o_{\text{fail}} \leftarrow \text{CaptureCurrentObservation}()$
        \State $P \leftarrow \pi_{\text{planner}}.\text{Replan}(I_{\text{user}}, P, k, o_{\text{fail}})$ 
        \State $o^{(k)}_{\text{prev}} \leftarrow o_{\text{fail}}$ 
    \EndIf
\EndWhile
\State \Return \texttt{Success}

\vspace{1em}
\Function{MonitoredExecute}{$s, o_{\text{start}}$}
    \State \texttt{is\_done} $\leftarrow$ \texttt{False}
    \State Start asynchronous execution of $s$ by $\pi_{\text{control}}$
    \While{not \texttt{is\_done} and not \texttt{Timeout}}
        \State $o_{\text{now}} \leftarrow \text{CaptureCurrentObservation}()$
        \State \texttt{is\_done} $\leftarrow \pi_{\text{planner}}.\text{Verify}(s, o_{\text{start}}, o_{\text{now}})$
        \State \texttt{wait}($\Delta t$) 
    \EndWhile
    \State Stop execution
    \State \Return \texttt{is\_done}
\EndFunction

\end{algorithmic}
\end{algorithm}

\section{Experiment}
\label{sec:experiment}
\subsection{Settings}
\subsubsection{Implementation of Low-level controller}
In this paper, we utilize the diffusion policy\cite{chi2023dp} with an additional frozen language encoder to generate action sequence. The language encoder, which is kept frozen during training, processes the natural language instruction and provides a semantically rich embedding. Meanwhile, the vision encoder extracts visual features from the current observation image. These language and observation embeddings are then combined using Feature-wise Linear Modulation (FiLM) conditioning\cite{perez2018film}, which allows the language features to modulate the visual features at each layer of the diffusion policy network. The training data for the low-level controller was collected from human demonstrations using the UMI gripper, and end-effector pose trajectories were generated via UMI's SLAM pipeline\cite{chi2024umi}.

\subsubsection{Training Details}
We implement GRPO optimization on eight H100 GPUs (80 GB), with one training epoch requiring approximately 28 hours. We fine-tune Qwen2.5-VL-7B\cite{bai2025qwen2} on the combined LEAP-L, LEAP-U, EgoPlan\cite{chen2023egoplan}, ShareRobot\cite{ji2025robobrain} and ERQA\cite{team2025gemini} with planning data. After careful cleaning, 57,000 triplets are used for training. Hyper-parameters are fixed across scenes: 3 epochs, per-device batch 8, gradient-accumulation 12, learning rate 1e-5 with cosine decay, $\beta$ = 0.04 for KL regularisation. For each prompt we sample 8 completions, truncate at 2048 tokens. 

\subsubsection{Benchmark}
To provide a comprehensive evaluation of RoboFarseer’s planning capabilities, we employ five benchmarks spanning multiple-choice question (MCQ) and open-ended question. LEAP-L and LEAP-U are derived from real UMI demonstrations, foregrounding temporal dependencies and cross-scene generalization. ShareRobot-Planning contributes open-vocabulary, fine-grained daily instructions harvested from the large-scale, crowd-annotated ShareRobot corpus\cite{ji2025robobrain}. EgoPlan-Bench2\cite{qiu2024egoplan2} focus on first-person videos, probing long-range decision-making under real human viewpoints. Note that the training set uses EgoPlan-IT, which is based on Epic-Kitchens\cite{chen2023egoplan}, whereas the evaluation set uses EgoPlan-bench2, which is based on Ego4D; thus, training and testing are performed on two disjoint datasets. RoboVQA furnishes a large-scale multimodal benchmark that couples open-vocabulary planning and verifying task-completion\cite{sermanet2024robovqa}. Finally, we also convert the test set of LEAP into MCQ format which comprises 655 data points, the scenario and instruction is different from training subset.

\subsubsection{Baselines}
We compare REVER with state-of-the-art VLMs across three groups:
\begin{itemize}
    \item commercial models: Gemini-2.5-Pro\cite{google2025gemini2p5pro}, Gemini-2.5-Flash\cite{google2025gemini2p5pro}, GPT-4.1\cite{openai2025gpt41}, GPT-4o\cite{hurst2024gpt} and Claude-Sonnet-4\cite{anthropic2025claude4}.
    \item Open-source models: Qwen2.5-VL-72B-Instruct\cite{bai2025qwen2} and Qwen2.5-VL-32B-Instruct\cite{bai2025qwen2}.
    \item Embodied-specific models: RoboBrain2-7B\cite{team2025robobrain2} and RoboBrain2-32B\cite{team2025robobrain2}.
\end{itemize}

Most embodied-specific models emphasize spatial reasoning rather than long-horizon planning\cite{yuan2025embodiedr1,azzolini2025cosmos,yuan2025fsd}. RoboBrain\cite{team2025robobrain2}, however, is explicitly designed for both spatial and temporal reasoning, so we adopt its 7B and 32B variants as our primary baselines.

\subsection{Planning Evaluation}
\begin{figure}[htbp]
    \centering
    \includegraphics[width=\linewidth]{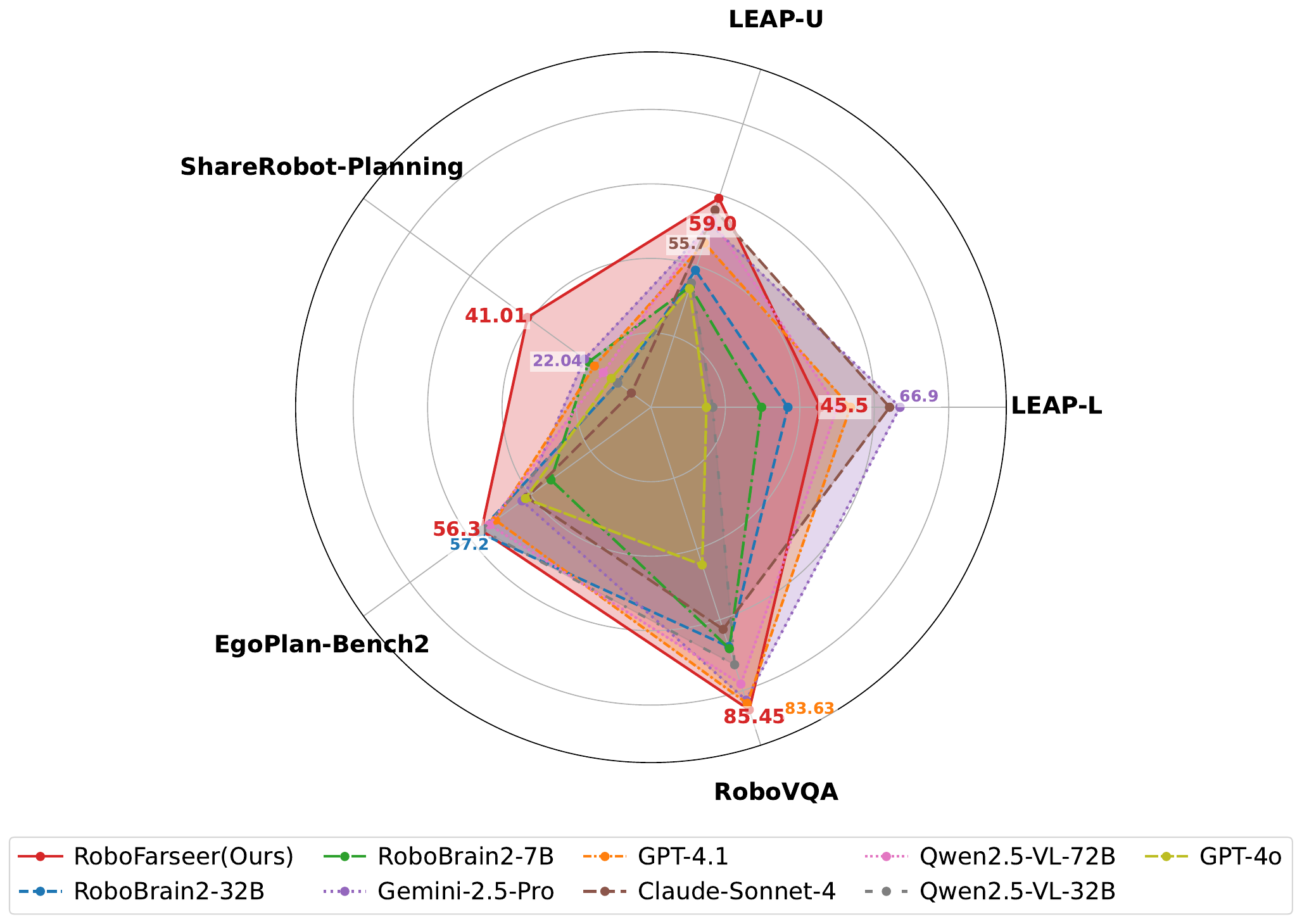}
    \caption{Planning accuracy (\%) on planning benchmarks.}
    \label{fig:radar}
\end{figure}
Fig. ~\ref{fig:radar} reports top-1 accuracy on the planning benchmarks.
On LEAP-L(MCQ), Gemini-2.5-Pro leads at 66.9 \%, followed by Claude-Sonnet-4 (64.1 \%).
RoboFarseer-7B scores 59.3 \%, outperforming every open-source competitor by at least 9 pp (Qwen2.5-VL-72B 50.1 \%, RoboBrain2-32B 36.8 \%, RoboBrain2-7B 29.7 \%) despite a 10× smaller parameter count.
On LEAP-U(MCQ), RoboVQA and ShareRobot-Planning, RoboFarseer obtains the highest mark among all models.
On EgoPlan2, RoboFarseer ranks second (56.4 \%) only to RoboBrain2-32B (57.2 \%), confirming that a 7B model trained with REVER already rivals 32B embodied specialists.
The strong scores of proprietary models validate the discriminative difficulty of our distractors and dataset quality; nevertheless, RoboFarseer remains within 6–8 pp of the best performer without any contrastive ranking objective, demonstrating parameter-efficient, generative planning that scales gracefully to larger backbones.
% \begin{table}[htbp]
%   \centering
%   \caption{Planning accuracy (\%) on MCQ planning benchmarks.}
%   \label{tab:mcq}
%   \begin{tabular}{lccc}
%     \toprule
%     Model & LEAP-L-MCQ & LEAP-U-MCQ & EgoPlan2\\
%     \midrule
%     RoboBrain2-7B & 29.7 & 34.1 & 33.2\\
%     RoboBrain2-32B & 36.7 & 38.7 & \textbf{57.2}\\
%     \midrule
%     GPT-4o & 14.9 & 33.5 & 41.7\\
%     GPT-4.1 & 53.6 & 46.4 & 51.6\\
%     Gemini-2.5-Flash & 24.6 & 25.8 & 37.0\\
%     Gemini-2.5-Pro & \textbf{66.9} & 51.7 & 42.8\\
%     Claude-Sonnet-4 & 64.1 & 55.7 & 41.26\\
%     \midrule
%     Qwen2.5-VL-32B & 16.6 & 35.1 & 56.2\\
%     Qwen2.5-VL-72B & 50.1 & 53.0 & 53.7\\
%     \midrule
%     \textbf{RoboFarseer}(7B) & 46.3 & \textbf{59.3} & 56.4\\
%     \bottomrule
%   \end{tabular}
% \end{table}
\begin{figure}[htbp]
    \centering
    \includegraphics[width=0.9\linewidth]{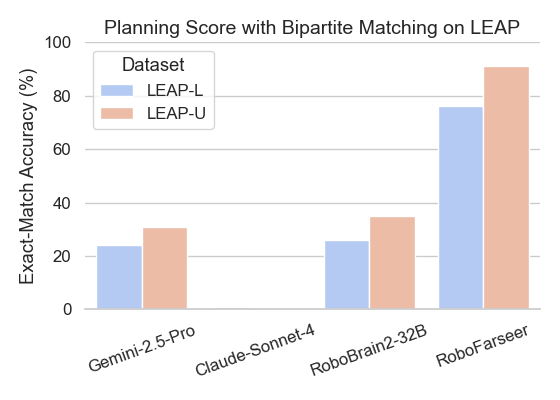}
    \caption{Planning score with bipartite matching on LEAP open-planning test set.}
    \label{fig:leap}
    \vspace{-3mm}
\end{figure}

We select three stronger competitors (Claude-Sonnet-4, Gemini-2.5-pro, RoboBrain2-32B), prompt them to generate complete plans. 
When models must generate complete plans, RoboFarseer establishes a clear margin (Fig. ~\ref{fig:leap}).
On LEAP-L, RoboFarseer achieves 76 \%, more than doubling the score of the strongest proprietary model, Gemini-2.5-Pro (24 \%), and tripling RoboBrain2-32B (26 \%).
Claude-Sonnet-4 obtains near-zero score because its output deviates from the required skill grammar.
An identical trend holds on LEAP-U.
The 45–60,pp indicate that the verifiable-reward fine-tuning in REVER successfully teaches the VLM to produce long, executable sequences, whereas frontier models fine-tuned on general dialogue or short-horizon data fail to maintain temporal coherence and grammar compliance in open-world manipulation planning.

\begin{figure}[htbp]
    \centering
    \includegraphics[width=0.9\linewidth]{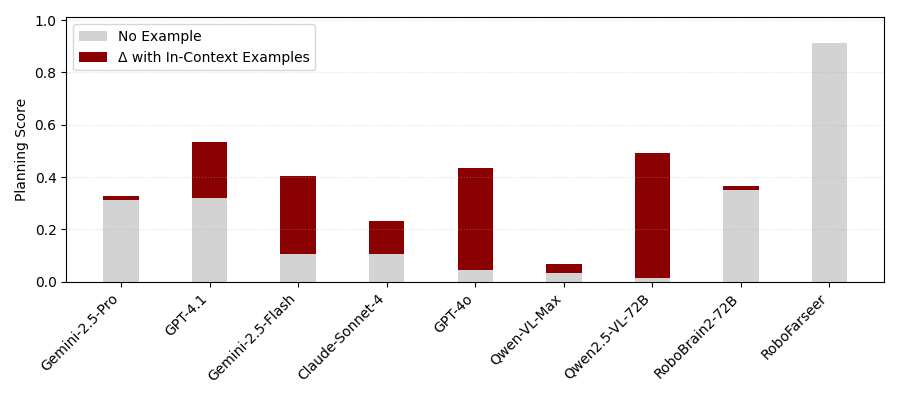}
    \caption{Impact of in-context examples on LEAP-U open planning.}
    \label{fig:incontext}
    \vspace{-3mm}
\end{figure}
To investigate the sharp performance drop of baselines on open-ended planning, we examined their adherence to the elementary-skill format specified in the prompt. As shown in Fig. \ref{fig:incontext}, supplying two in-context examples in the prompt consistently improved planning performance than zero-shot planning. RoboFarseer maintained 92 \% exact-match, confirming that the verifiable-reward training in REVER enforces format fidelity beyond simple few-shot prompting.

\subsection{Real-World Evaluation}
\label{real-world}
\begin{figure}[htbp]
  \centering
  \includegraphics[width=0.24\linewidth]{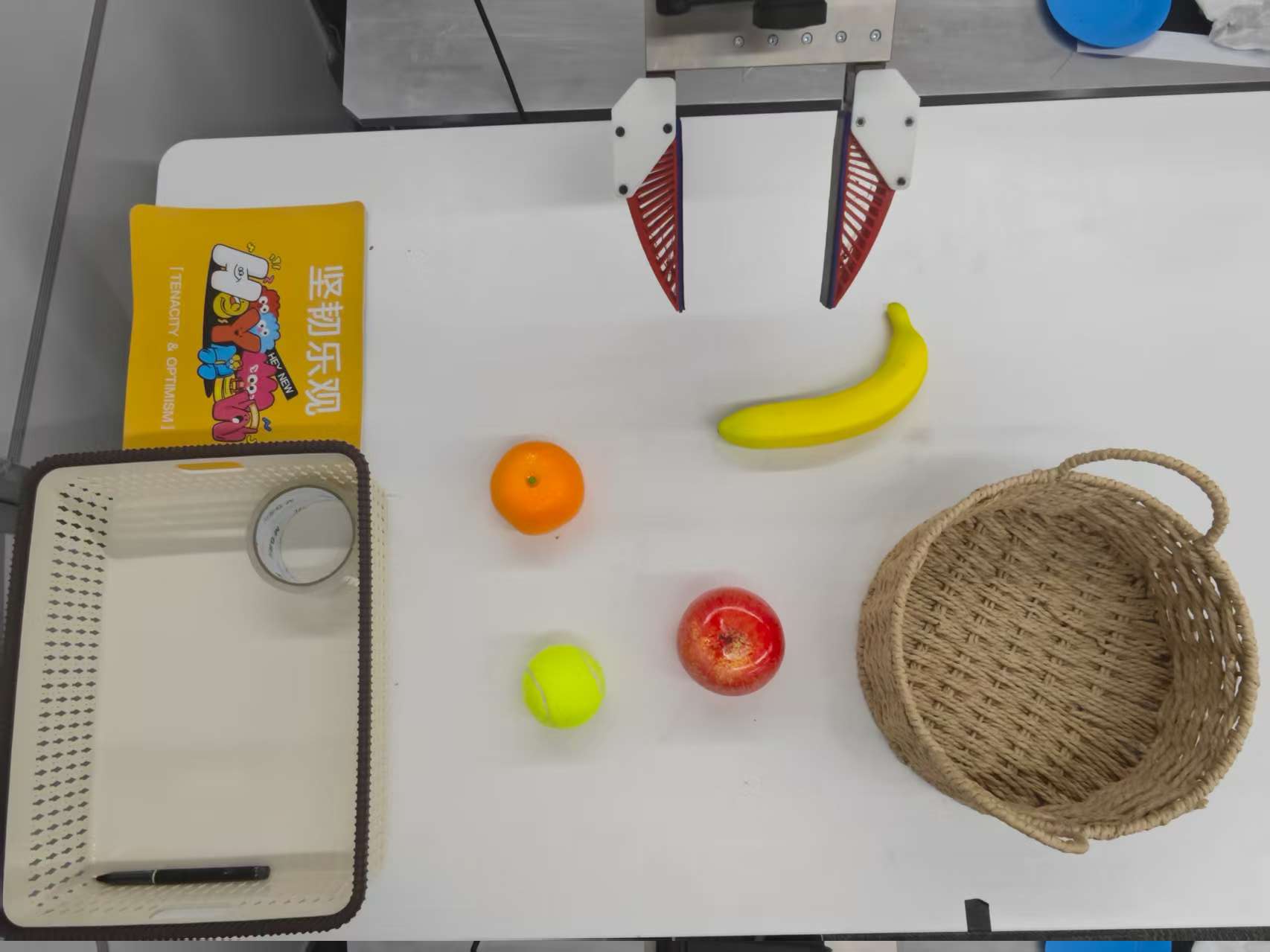}\hfill
  \includegraphics[width=0.24\linewidth]{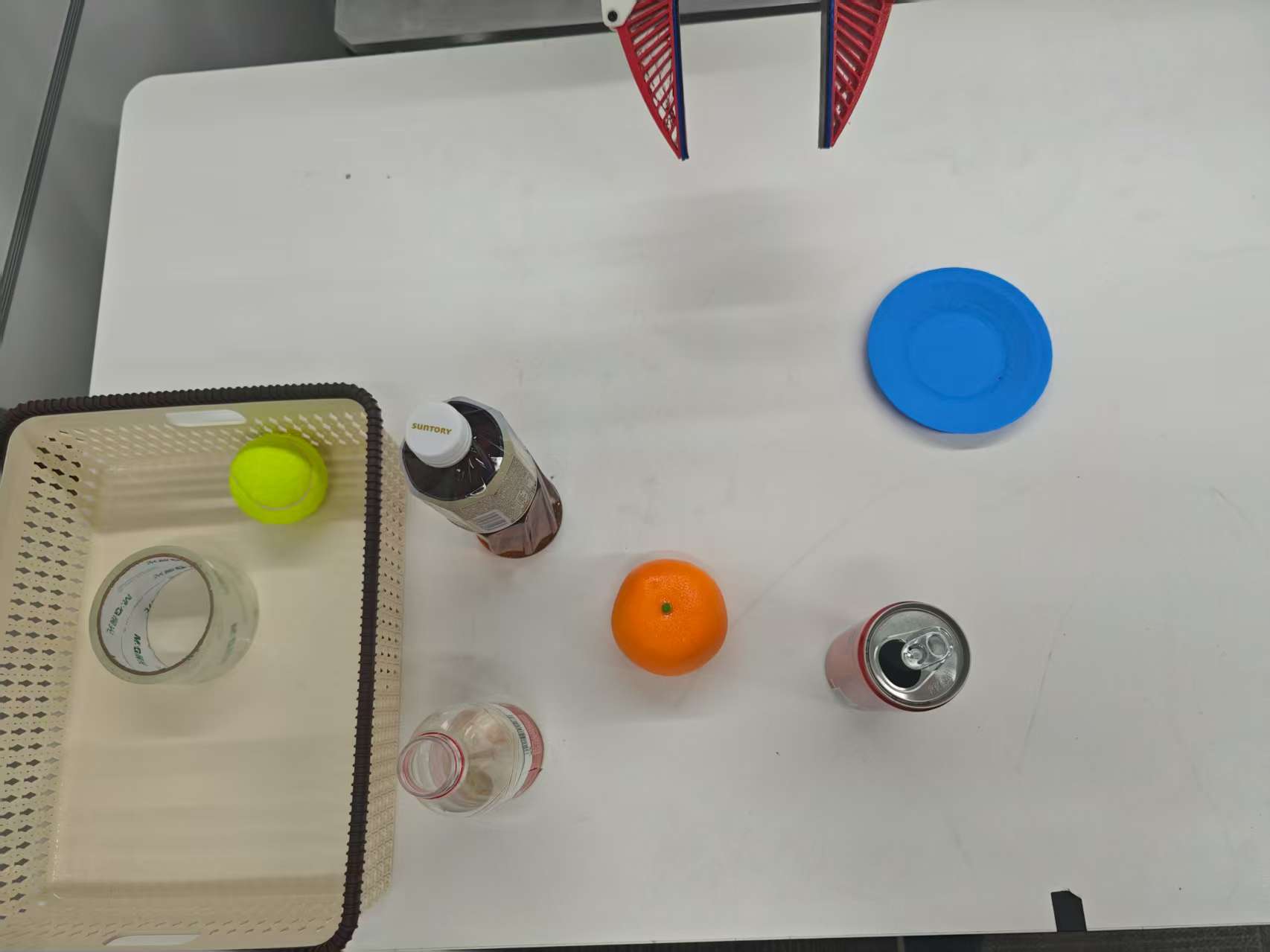}\hfill
  \includegraphics[width=0.24\linewidth]{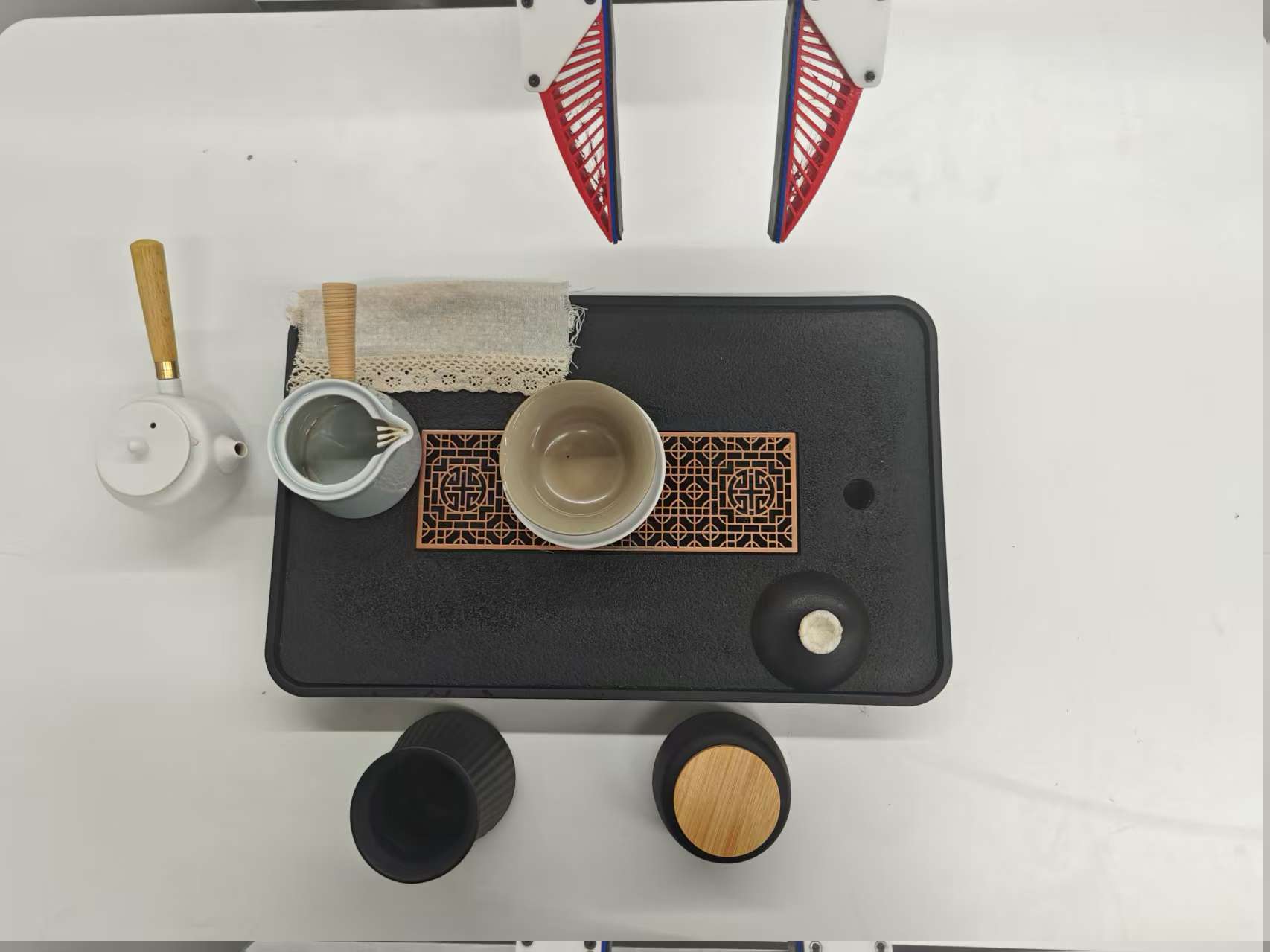}\hfill
  \includegraphics[width=0.24\linewidth]{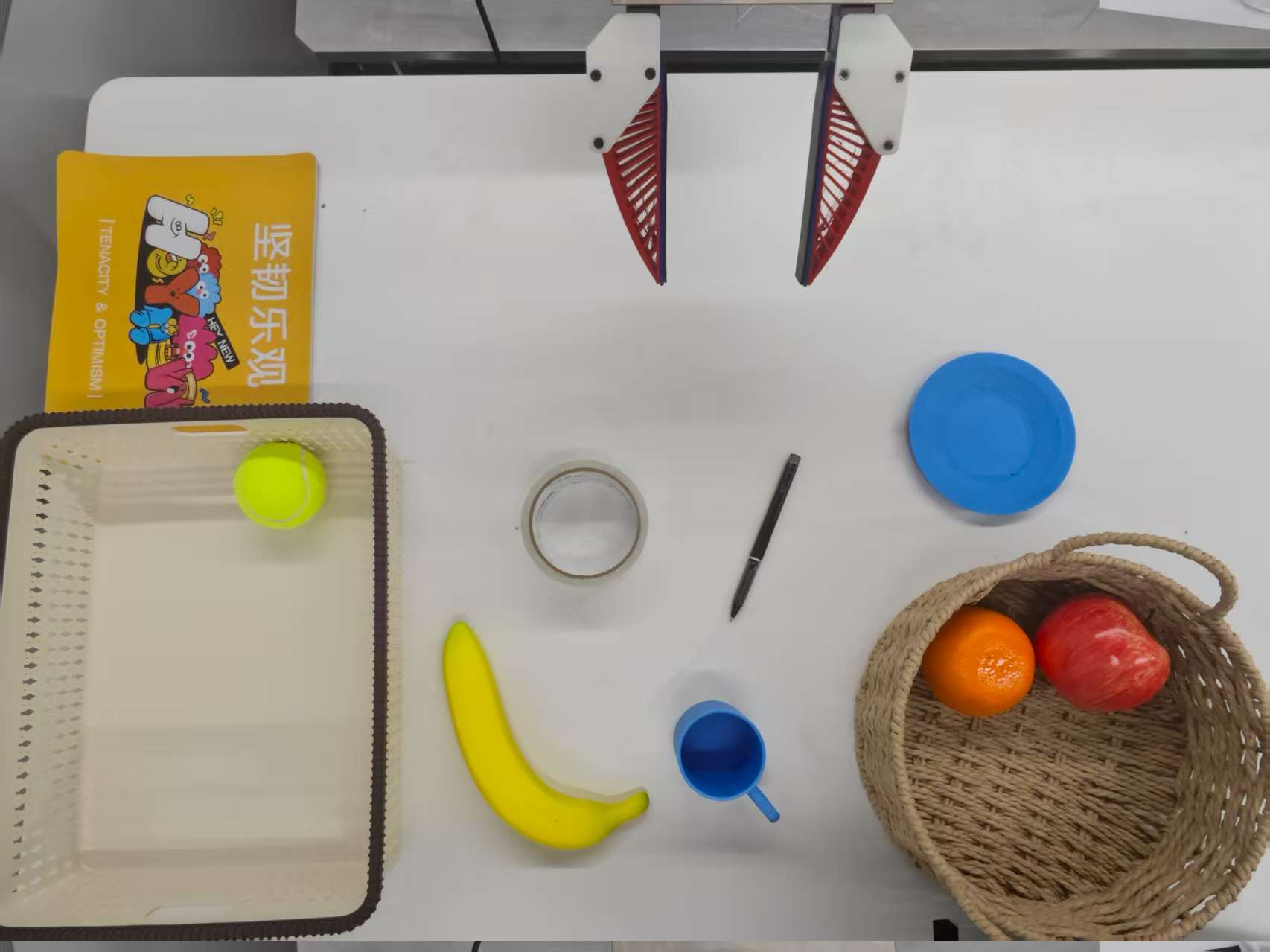}
  \caption{Real-World Evaluation Scenarios}
  \label{fig:Real-World}
  \vspace{-3mm}
\end{figure}
\label{sec:real_world_exp}

\begin{table*}[htbp]
\centering
\caption{
  Quantitative results on 10 instructions.  
  Each instruction was evaluated in 10 episodes with different initial scenes.  
  “Overall success” = percentage of fully completed tasks;  
  “Step success” = percentage of sub-tasks completed.  
  The right-most column gives the Overall success of an ablation that removes the VLM planner and uses a control policy directly.
}
\label{tab:abstract_instructions}
\renewcommand{\arraystretch}{1.1}
\small
\begin{tabular}{@{}p{2.8cm}p{6.9cm}ccc@{}}
\toprule
\textbf{ } & \textbf{Instruction} & \textbf{Overall success} & \textbf{Step success} & \textbf{w/o VLM planner} \\
\midrule
\multirow{3}{*}{Tidy up Desktop}
 & 1. Tidy the small objects into their containers & 80\% & 91.4\% & 0\% \\
 & 2. Put all the fruits into the basket & 80\% & 94.3\% & 40\% \\
 & 3. Put all round objects into the basket & 90\% & 96.9\% & 30\% \\
 & 4. Organize all stationery (pens and tapes) & 60\% & 77.8\% & 0\% \\
\midrule
\multirow{2}{*}{Brew tea}
 & 5. Make me a cup of tea & 70\% & 83.9\% & 0\% \\
 & 6. Make tea and cover the cup with the lid & 50\% & 61.3\% & 0\% \\
\midrule
\multirow{4}{*}{Bring food \& drinks}
 & 7. I’m hungry, bring me something to eat & 100\% & 100\% &10\% \\
 & 8. I’m thirsty, bring me some drinks & 90\% & 95.0\% & 20\% \\
 & 9. I don’t like sweet drinks, bring me some drinks & 70\% & 80.0\% & 0\% \\
 & 10. I want carbonated drinks & 70\% & 85.0\% & 10\% \\
\bottomrule
\end{tabular}
\end{table*}
We evaluate REVER on a Dobot Nova5 6-DoF robot arm with GoPro10 camera in three household scenes: \textit{Tidy up Desktop }, \textit{Brew Tea}, and \textit{Bring Food \& Drinks}.
Each scene is set up only once; objects are placed in random poses and location. Below we describe the three scenarios:
\paragraph{Tidy up Desktop}
The  contains a mix of fruits, cup, and assorted daily items together with several containers (basket, box, mouse pad, tray).
Commands range from category-level (“put all fruits into the basket”) to geometry-based (“gather every round object”) and require the robot to segment the scene, identify relevant instances, and execute a variable-length pick-place sequence while avoiding collisions with irrelevant objects.
\paragraph{Brew Tea}
This scene highlights long-horizon and complex manipulation.
A teapot, a dispenser, a teacup, and a lid are randomly arranged on a tea tray.
The instruction may triggers a multi-stage plan:
(1) Open the teacup lid;
(2) decant water from the teapot into the fairness pitcher;
(3) pour the infused tea from the pitcher into the teacup;
(4) place the lid onto the cup.

\paragraph{Bring Food \& Drinks}
User requests include high-level desires (“I’m hungry, bring me something to eat”), attribute constraints (“I don’t like sweet drinks, bring me some drink”).
The robot must ground the utterance to an object that satisfies nutritional and preference constraints.

We conducted experiments for each of ten abstract user instructions, running ten episodes per instruction; the results are reported in Table~\ref{tab:abstract_instructions}. None of these exact instructions appear in the training data. Instructions 1, 5 and 6 have close paraphrastic variants in the training set, whereas the remaining seven instructions are completely unseen during training. 

First, the “Bring food \& drinks” family achieves the highest average overall success rate (90 \%), with step-level accuracies $\ge$ 95 \%, indicating that unambiguous semantic anchors and short action chains are easy for the planner to align.
Second, within “Tidy up Desktop”, shape-based grouping reaches 90 \%/96.9 \%, whereas the finer-grained stationery arrangement drops to 60 \%/77.8 \%. The latter suffers from thin, low-contrast objects (slender pens and transparent tape) that are difficult to detect and grasp, demonstrating that both geometric and material properties compound spatial–category complexity.
Third, “Brew tea” emerges as the most challenging category: pouring demands millimetre-level precision, the liquid surface is hard to localise from a single RGB view, and the small, slippery cup lid resists stable grasping. These factors jointly reduce overall success by 10–20 percentage points and lower step success, confirming that early perception or manipulation failures in longer horizons trigger cascading errors.
Comparing the ablation without the VLM planner (right-most column), the pure end-to-end control policy averages only 14 \% overall success and attains 0 \% on the tea tasks, corroborating that explicit high-level planning is indispensable for compositional, long-horizon manipulation.

\begin{figure}[htbp]
    \centering
    \includegraphics[width=\linewidth,trim=2 2 2 2,clip]{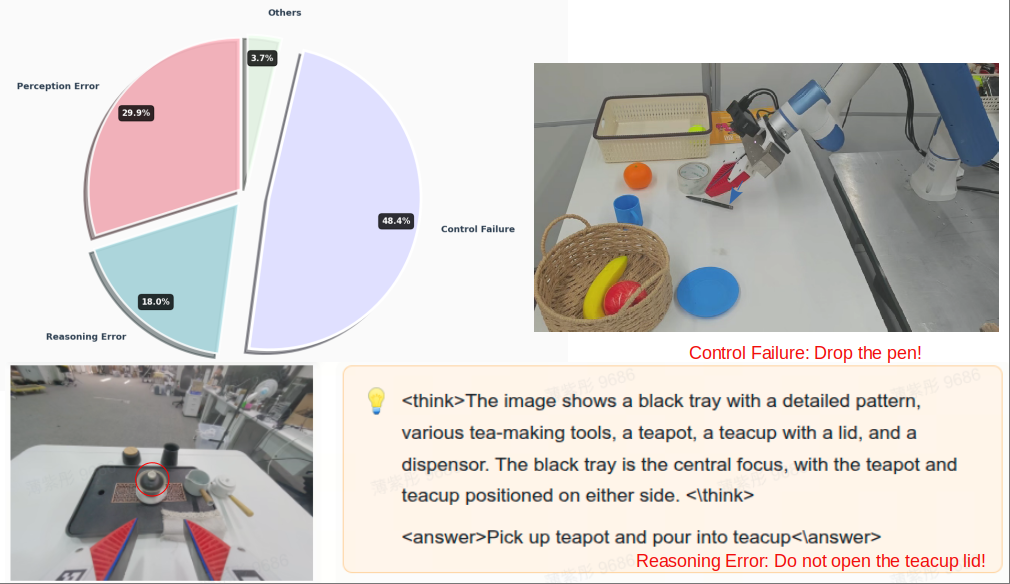}
    \caption{Failure cause distribution and examples.}
    \label{fig:failure}
    \vspace{-3mm}
\end{figure}
We examined every unsuccessful case that did not reach the instructed goal, as shown in Fig. ~\ref{fig:failure}. Perception errors—where the planner mis-classified object state or pose—accounted for 26.9\% of these failures. Reasoning errors, i.e., incorrect decomposition of the high-level instruction into elementary skills, comprised 16.2\%. The dominant source (43.6\%) was downstream control failure: although the generated plan was correct, the low-level policy could not achieve the required gripper pose, insertion angle, or pouring motion. The remaining 3.3\% were miscellaneous issues such as malformed output syntax. These statistics indicate that the current bottleneck is not planning but execution, suggesting that future work should prioritise more robust visuo-motor control or real-time reactive refinement of low-level actions.

\section{Conclusion}
\label{sec:conclusion}

We introduce REVER, a training framework that equips a VLM with a verifiable, grammar-aware reward to perform reinforced, long-horizon embodied planning.
To make learning feasible, we collected large-scale kinesthetic demonstrations via the UMI interface and released two complementary datasets, LEAP-L and LEAP-U.
The resulting model, RoboFarseer, emits interpretable chain-of-thought and simultaneously acts as an open-loop planner and a closed-loop progress monitor.
Quantitatively, RoboFarseer matches or surpasses the performance of proprietary models that are orders of magnitude larger on seven planning benchmarks, while on open-ended planning it exceeds the strongest baseline by more than 40\%. In real-world, long-horizon tasks, the complete system boosts overall success by roughly 60\% compared with the low-level controller without the planner.
All model weights, datasets and source code are made publicly available to accelerate future embodied-AI research.
In future work, we will enlarge the skill vocabulary, incorporate reactive re-planning mechanisms, and extend the framework to bimanual and loco-manipulation scenarios.
\bibliographystyle{IEEEtran}
%\bibliography{./bibliography/IEEEabrv,./bibliography/IEEEexample}
%\bibliographystyle{IEEEtran}
\bibliography{MyBibliography}

% Generated by IEEEtran.bst, version: 1.12 (2007/01/11)
\begin{thebibliography}{10}
\providecommand{\url}[1]{#1}
\csname url@samestyle\endcsname
\providecommand{\newblock}{\relax}
\providecommand{\bibinfo}[2]{#2}
\providecommand{\BIBentrySTDinterwordspacing}{\spaceskip=0pt\relax}
\providecommand{\BIBentryALTinterwordstretchfactor}{4}
\providecommand{\BIBentryALTinterwordspacing}{\spaceskip=\fontdimen2\font plus
\BIBentryALTinterwordstretchfactor\fontdimen3\font minus \fontdimen4\font\relax}
\providecommand{\BIBforeignlanguage}[2]{{%
\expandafter\ifx\csname l@#1\endcsname\relax
\typeout{** WARNING: IEEEtran.bst: No hyphenation pattern has been}%
\typeout{** loaded for the language `#1'. Using the pattern for}%
\typeout{** the default language instead.}%
\else
\language=\csname l@#1\endcsname
\fi
#2}}
\providecommand{\BIBdecl}{\relax}
\BIBdecl

\bibitem{wu2023embodied}
Z.~Wu, Z.~Wang, X.~Xu, J.~Lu, and H.~Yan, ``Embodied task planning with large language models,'' \emph{arXiv preprint arXiv:2307.01848}, 2023.

\bibitem{shi2025hirobot}
L.~X. Shi, B.~Ichter, M.~Equi, L.~Ke, K.~Pertsch, Q.~Vuong, J.~Tanner, A.~Walling, H.~Wang, N.~Fusai \emph{et~al.}, ``Hi robot: Open-ended instruction following with hierarchical vision-language-action models,'' \emph{arXiv preprint arXiv:2502.19417}, 2025.

\bibitem{chi2023dp}
C.~Chi, Z.~Xu, S.~Feng, E.~Cousineau, Y.~Du, B.~Burchfiel, R.~Tedrake, and S.~Song, ``Diffusion policy: Visuomotor policy learning via action diffusion,'' \emph{arXiv preprint arXiv:2303.04137}, 2023.

\bibitem{zitkovich2023rt}
B.~Zitkovich, T.~Yu, S.~Xu, P.~Xu, T.~Xiao, F.~Xia, J.~Wu, P.~Wohlhart, S.~Welker, A.~Wahid \emph{et~al.}, ``Rt-2: Vision-language-action models transfer web knowledge to robotic control,'' in \emph{Conference on Robot Learning}.\hskip 1em plus 0.5em minus 0.4em\relax PMLR, 2023, pp. 2165--2183.

\bibitem{bordes2024introduction}
F.~Bordes, R.~Y. Pang, A.~Ajay, A.~C. Li, A.~Bardes, S.~Petryk, O.~Ma{\~n}as, Z.~Lin, A.~Mahmoud, B.~Jayaraman \emph{et~al.}, ``An introduction to vision-language modeling,'' \emph{arXiv preprint arXiv:2405.17247}, 2024.

\bibitem{huang2025vlm}
Z.~Huang, Z.~Sheng, Y.~Qu, J.~You, and S.~Chen, ``Vlm-rl: A unified vision language models and reinforcement learning framework for safe autonomous driving,'' \emph{Transportation Research Part C: Emerging Technologies}, vol. 180, p. 105321, 2025.

\bibitem{wu2025reinforced}
D.~Wu, J.~Fan, J.~Zang, G.~Wang, W.~Yin, W.~Li, and B.~Jin, ``Reinforced reasoning for embodied planning,'' \emph{arXiv preprint arXiv:2505.22050}, 2025.

\bibitem{yuan2025embodiedr1}
Y.~Yuan, H.~Cui, Y.~Huang, Y.~Chen, F.~Ni, Z.~Dong, P.~Li, Y.~Zheng, and J.~Hao, ``Embodied-r1: Reinforced embodied reasoning for general robotic manipulation,'' \emph{arXiv preprint arXiv:2508.13998}, 2025.

\bibitem{azzolini2025cosmos}
A.~Azzolini, J.~Bai, H.~Brandon, J.~Cao, P.~Chattopadhyay, H.~Chen, J.~Chu, Y.~Cui, J.~Diamond, Y.~Ding \emph{et~al.}, ``Cosmos-reason1: From physical common sense to embodied reasoning,'' \emph{arXiv preprint arXiv:2503.15558}, 2025.

\bibitem{ji2025robobrain}
Y.~Ji, H.~Tan, J.~Shi, X.~Hao, Y.~Zhang, H.~Zhang, P.~Wang, M.~Zhao, Y.~Mu, P.~An \emph{et~al.}, ``Robobrain: A unified brain model for robotic manipulation from abstract to concrete,'' in \emph{Proceedings of the Computer Vision and Pattern Recognition Conference}, 2025, pp. 1724--1734.

\bibitem{whitehouse2025j1}
C.~Whitehouse, T.~Wang, P.~Yu, X.~Li, J.~Weston, I.~Kulikov, and S.~Saha, ``J1: Incentivizing thinking in llm-as-a-judge via reinforcement learning,'' \emph{arXiv preprint arXiv:2505.10320}, 2025.

\bibitem{chi2024umi}
C.~Chi, Z.~Xu, C.~Pan, E.~Cousineau, B.~Burchfiel, S.~Feng, R.~Tedrake, and S.~Song, ``Universal manipulation interface: In-the-wild robot teaching without in-the-wild robots,'' \emph{arXiv preprint arXiv:2402.10329}, 2024.

\bibitem{guo2025deepseek}
D.~Guo, D.~Yang, H.~Zhang, J.~Song, R.~Zhang, R.~Xu, Q.~Zhu, S.~Ma, P.~Wang, X.~Bi \emph{et~al.}, ``Deepseek-r1: Incentivizing reasoning capability in llms via reinforcement learning,'' \emph{arXiv preprint arXiv:2501.12948}, 2025.

\bibitem{bai2025qwen2}
S.~Bai, K.~Chen, X.~Liu, J.~Wang, W.~Ge, S.~Song, K.~Dang, P.~Wang, S.~Wang, J.~Tang \emph{et~al.}, ``Qwen2.5-vl technical report,'' \emph{arXiv preprint arXiv:2502.13923}, 2025.

\bibitem{shin2024socratic}
S.~Shin, S.~Jeon, J.~Kim, G.-C. Kang, and B.-T. Zhang, ``Socratic planner: Inquiry-based zero-shot planning for embodied instruction following,'' \emph{CoRR}, 2024.

\bibitem{fei2025unleashing}
Z.~Fei, L.~Ji, S.~Wang, J.~Shi, J.~Gong, and X.~Qiu, ``Unleashing embodied task planning ability in llms via reinforcement learning,'' \emph{arXiv preprint arXiv:2506.23127}, 2025.

\bibitem{driess2023palme}
D.~Driess, F.~Xia, M.~S. Sajjadi, C.~Lynch, A.~Chowdhery, B.~Ichter, A.~Wahid, J.~Tompson, Q.~Vuong, T.~Yu \emph{et~al.}, ``Palm-e: an embodied multimodal language model,'' in \emph{Proceedings of the 40th International Conference on Machine Learning}, 2023, pp. 8469--8488.

\bibitem{zhang2025embodiedreasoner}
W.~Zhang, M.~Wang, G.~Liu, X.~Huixin, Y.~Jiang, Y.~Shen, G.~Hou, Z.~Zheng, H.~Zhang, X.~Li \emph{et~al.}, ``Embodied-reasoner: Synergizing visual search, reasoning, and action for embodied interactive tasks,'' \emph{arXiv preprint arXiv:2503.21696}, 2025.

\bibitem{wen2025dexvla}
J.~Wen, Y.~Zhu, J.~Li, Z.~Tang, C.~Shen, and F.~Feng, ``Dexvla: Vision-language model with plug-in diffusion expert for general robot control,'' \emph{arXiv preprint arXiv:2502.05855}, 2025.

\bibitem{cai2025cookbench}
M.~Cai, X.~Chen, Y.~An, J.~Zhang, X.~Wang, W.~Xu, W.~Zhang, and T.~Liu, ``Cookbench: A long-horizon embodied planning benchmark for complex cooking scenarios,'' \emph{arXiv preprint arXiv:2508.03232}, 2025.

\bibitem{yang2025embodiedbench}
R.~Yang, H.~Chen, J.~Zhang, M.~Zhao, C.~Qian, K.~Wang, Q.~Wang, T.~V. Koripella, M.~Movahedi, M.~Li \emph{et~al.}, ``Embodiedbench: Comprehensive benchmarking multi-modal large language models for vision-driven embodied agents,'' \emph{arXiv preprint arXiv:2502.09560}, 2025.

\bibitem{shao2025large}
R.~Shao, W.~Li, L.~Zhang, R.~Zhang, Z.~Liu, R.~Chen, and L.~Nie, ``Large vlm-based vision-language-action models for robotic manipulation: A survey,'' \emph{arXiv preprint arXiv:2508.13073}, 2025.

\bibitem{chen2023robogpt}
Y.~Chen, W.~Cui, Y.~Chen, M.~Tan, X.~Zhang, D.~Zhao, and H.~Wang, ``Robogpt: an intelligent agent of making embodied long-term decisions for daily instruction tasks,'' \emph{arXiv preprint arXiv:2311.15649}, 2023.

\bibitem{rafailov2023dpo}
R.~Rafailov, A.~Sharma, E.~Mitchell, C.~D. Manning, S.~Ermon, and C.~Finn, ``Direct preference optimization: Your language model is secretly a reward model,'' \emph{Advances in neural information processing systems}, vol.~36, pp. 53\,728--53\,741, 2023.

\bibitem{shao2024deepseekmath}
Z.~Shao, P.~Wang, Q.~Zhu, R.~Xu, J.~Song, X.~Bi, H.~Zhang, M.~Zhang, Y.~Li \emph{et~al.}, ``Deepseekmath: Pushing the limits of mathematical reasoning in open language models,'' \emph{arXiv preprint arXiv:2402.03300}, 2024.

\bibitem{yuan2025fsd}
Y.~Yuan, H.~Cui, Y.~Chen, Z.~Dong, F.~Ni, L.~Kou, J.~Liu, P.~Li, Y.~Zheng, and J.~Hao, ``From seeing to doing: Bridging reasoning and decision for robotic manipulation,'' \emph{arXiv preprint arXiv:2505.08548}, 2025.

\bibitem{team2025robobrain2}
B.~R. Team, M.~Cao, H.~Tan, Y.~Ji, M.~Lin, Z.~Li, Z.~Cao, P.~Wang, E.~Zhou, Y.~Han \emph{et~al.}, ``Robobrain 2.0 technical report,'' \emph{arXiv preprint arXiv:2507.02029}, 2025.

\bibitem{brohan2022rt1}
A.~Brohan, N.~Brown, J.~Carbajal, Y.~Chebotar, J.~Dabis, C.~Finn, K.~Gopalakrishnan, K.~Hausman, A.~Herzog, J.~Hsu \emph{et~al.}, ``Rt-1: Robotics transformer for real-world control at scale,'' \emph{arXiv preprint arXiv:2212.06817}, 2022.

\bibitem{o2024oxe}
A.~O’Neill, A.~Rehman, A.~Maddukuri, A.~Gupta, A.~Padalkar, A.~Lee, A.~Pooley, A.~Gupta, A.~Mandlekar, A.~Jain \emph{et~al.}, ``Open x-embodiment: Robotic learning datasets and rt-x models: Open x-embodiment collaboration 0,'' in \emph{2024 IEEE International Conference on Robotics and Automation (ICRA)}.\hskip 1em plus 0.5em minus 0.4em\relax IEEE, 2024, pp. 6892--6903.

\bibitem{schulman2017ppo}
J.~Schulman, F.~Wolski, P.~Dhariwal, A.~Radford, and O.~Klimov, ``Proximal policy optimization algorithms,'' \emph{arXiv preprint arXiv:1707.06347}, 2017.

\bibitem{perez2018film}
E.~Perez, F.~Strub, H.~De~Vries, V.~Dumoulin, and A.~Courville, ``Film: Visual reasoning with a general conditioning layer,'' in \emph{Proceedings of the AAAI conference on artificial intelligence}, vol.~32, no.~1, 2018.

\bibitem{chen2023egoplan}
Y.~Chen, Y.~Ge, Y.~Ge, M.~Ding, B.~Li, R.~Wang, R.~Xu, Y.~Shan, and X.~Liu, ``Egoplan-bench: Benchmarking multimodal large language models for human-level planning,'' \emph{arXiv preprint arXiv:2312.06722}, 2023.

\bibitem{team2025gemini}
G.~R. Team, S.~Abeyruwan, J.~Ainslie, J.-B. Alayrac, M.~G. Arenas, T.~Armstrong, A.~Balakrishna, R.~Baruch, M.~Bauza, M.~Blokzijl \emph{et~al.}, ``Gemini robotics: Bringing ai into the physical world,'' \emph{arXiv preprint arXiv:2503.20020}, 2025.

\bibitem{qiu2024egoplan2}
L.~Qiu, Y.~Chen, Y.~Ge, Y.~Ge, Y.~Shan, and X.~Liu, ``Egoplan-bench2: A benchmark for multimodal large language model planning in real-world scenarios,'' \emph{arXiv preprint arXiv:2412.04447}, 2024.

\bibitem{sermanet2024robovqa}
P.~Sermanet, T.~Ding, J.~Zhao, F.~Xia, D.~Dwibedi, K.~Gopalakrishnan, C.~Chan, G.~Dulac-Arnold, S.~Maddineni, N.~J. Joshi \emph{et~al.}, ``Robovqa: Multimodal long-horizon reasoning for robotics,'' in \emph{2024 IEEE International Conference on Robotics and Automation (ICRA)}.\hskip 1em plus 0.5em minus 0.4em\relax IEEE, 2024, pp. 645--652.

\bibitem{google2025gemini2p5pro}
{Google}, ``Gemini 2.5 pro preview: Even better coding performance,'' \url{https://developers.googleblog.com/en/gemini-2-5-pro-io-improved-coding-performance/}, 2025, accessed: 2025-05-06.

\bibitem{openai2025gpt41}
{OpenAI}, ``Introducing {GPT-4.1} in the {API},'' \url{https://openai.com/index/gpt-4-1/}, 2025, accessed: 2025-07-28.

\bibitem{hurst2024gpt}
A.~Hurst, A.~Lerer, A.~P. Goucher, A.~Perelman, A.~Ramesh, A.~Clark, A.~Ostrow, A.~Welihinda, A.~Hayes, A.~Radford \emph{et~al.}, ``Gpt-4o system card,'' \emph{arXiv preprint arXiv:2410.21276}, 2024.

\bibitem{anthropic2025claude4}
{Anthropic}, ``Introducing {Claude} 4,'' \url{https://www.anthropic.com/news/claude-4}, 2025, accessed: 2025-07-28.

\end{thebibliography}

% In your Appendix section
\appendix
% Use figure* environment to span both columns
\begin{figure*}[htbp]
    \begin{promptbox}{Planning Task Prompt Example}
    \begin{lstlisting}
    <image> You are a helpful and meticulous robot assistant. Your goal is to help users with real-world tasks using your gripper.
    
    Your available skills are:
    - Put [object] on [location].
    - Put [object] into [location].
    - Pick up [object] and pour into [location].
    - Pick up [object].
    - Open [object].
    - Push [object].
    - Pour into [location].
    - Place on [location].
    - Place into [location].
    
    The user request is: Tidy up the small items on the desktop
    
    Based on the image, describe what you see and generate a step-by-step plan to fulfill the users request. You must ONLY use the available skills listed above. Each step in your plan must exactly match one of the skill formats.
    
    The plan should be numbered like this:
    1. [Skill with object and location]
    2. [Skill with object and location]
    ...
    
    Avoid empty, duplicate, or irrelevant steps.
    
    \end{lstlisting}
    \end{promptbox}
\end{figure*}

\begin{figure*}[htbp]
    \begin{promptbox}{Completion Verification Prompt Example}
    \begin{lstlisting}
    <image><image>You are a precise robot assistant tasked with verifying if an action has been successfully completed.
    
    The first image shows the initial state of the environment before the action. The second image shows the final state after the action was attempted.
    
    Based on your observation of both images, determine if the following action was completed:
    
    # Pick up the teacup and place it on the saucer. #
    
    Output ONLY True if the second image clearly shows the object is in the target location as described in the action. Otherwise, output ONLY False.
    
    \end{lstlisting}
    \end{promptbox}
\end{figure*}

\end{document}